\documentclass[letterpaper, 10 pt, conference]{ieeeconf}  

\IEEEoverridecommandlockouts                              

\usepackage{graphicx} 
\usepackage{amsmath} 
\usepackage{amssymb}  
\usepackage{subfigure}

\title{\LARGE \bf
Position prediction using disturbance observer for planar pushing*
}

\author{Jongrae Kim$^{1}$
\thanks{*This work was supported by the EPSRC Research Grant, EP/N010523/1, 
Balancing the impact of city infrastructure engineering on natural systems
using robots.}
\thanks{$^{1}$Jongrae Kim is with Institute of Design, Robotics \& Optimisation (iDRO),
	School of Mechanical Engineering, University of Leeds, Leeds LS1 9JT, the UK,
        {\tt\small menjkim@leeds.ac.uk}}%
}

\begin{document}

\maketitle
\thispagestyle{empty}
\pagestyle{empty}

\begin{abstract}

The position and the orientation of a rigid body object pushed by a robot on a planar 
surface are extremely difficult to predict.
In this paper, the prediction problem is formulated as a disturbance observer design problem.
The disturbance observer provides accurate estimation of the total sum of
model errors and external disturbances acting on the object. 
From the estimation results, it is revealed that
there is a strong linear relationship between the applied force or torque
and the estimated disturbances. 
The proposed prediction algorithm has two phases: the identification \& the prediction. 
During the identification phase, the linear relationship is identified from
the observer output using a recursive least-square algorithm. In the prediction phase,
the identified linear relationship is used with a force plan, which would be
provided by a mission planner, to predict the position and the orientation
of an object. The algorithm is tested for six different push experimental data 
available from the MIT MCube Lab. The proposed algorithm shows improved performance
in reducing the prediction error compared to a simple correction algorithm.
\end{abstract}

\section{Introduction}

The position and the orientation of a rigid body object pushed by a robot on a planar 
surface are extremely difficult to predict.
The stochastic nature of physical interactions, \cite{7989345}, between objects, surfaces
and robots makes an accurate prediction of the position and orientation extraordinary challenging.
There are abundant researches on this topic in the past and the current 
and a few directly related ones to the problems studied in this paper are mentioned in the following. 

A nonlinear optimisation problem is solved in \cite{doi:10.1177/0278364917698749}
to identify the inertial parameters and the contact forces of multiple rigid bodies
in a plane with unknown frictional forces, 
while the contact geometries are assumed to be known. 
A hybrid model combined an analytical model with a convolutional neural network is presented 
in \cite{Kloss2017CombiningLA} for predicting the transnational and
rotational motions of an object in a planar pushing, where the position and velocity information of
the pusher is used. One of the crucial assumption of the algorithm is a quasi-steady state of 
the motion. Its prediction performance is degraded for different shape objects or high push velocities.
In \cite{Yu2018RealtimeSE}, a state estimation problem for a planar object motion is 
solved using visual and tactile sensor measurements. This algorithm, however, does not have 
any prediction part.

These studies focused on understanding of the push dynamics, combining
a dynamic model with the machine learning algorithm, or applying purely the machine learning
algorithm. In this paper, it is formulated as a disturbance observer design problem.
Disturbance observer is one of the widely used control methods. It is mainly used 
to compensate the influences of external disturbances and parametric uncertainties in 
a system to achieve robustness towards those perturbations and recover a desired performance.
A survey for various disturbance observers is found in \cite{7265050}.

A disturbance observer called Q-filter is adopted in this paper, which was first 
presented in \cite{4158825}. The first Q-filter observer is
used to estimate disturbances and uncertainties in a DC motor control. It showed promising
performances becoming one of the most popular methods for compensating disturbance effects.
Stability and robustness of the Q-filter based disturbance observer is presented 
in \cite{Shim2016}. 
It is not possible, however, that the disturbance observer is simply deployed for the push
prediction problem as the observer provides only the current estimation. A prediction
algorithm based on the estimated disturbances is required.

This paper is organised as follows: firstly, the push data from the MIT MCube lab is introduced 
\cite{MCubeLab_PushData}, six test scenarios are chosen, and the total force applied is estimated; 
secondly, a disturbance observer is designed for estimating unknown forces,
a prediction algorithm for position \& orientation prediction is presented,
and the performance of the prediction algorithm is demonstrated
using the test scenarios; and, finally, discussions \& future work are presented.

\section{Push Data \& Pre-Processing}
In this section, the push data available from the MIT MCube lab is introduced and pre-processed. 
The equation of motions for the translation
and the rotation are derived, and the total force and torque applied to an object are identified.

\subsection{Push Data}
The MIT MCube lab performed planar push experiments using a robot arm for various shape of objects,
types of surfaces, while the position and the orientation of the pusher, 
the force experienced in the tip of the pusher, and the position and the orientation of the objects 
are recorded with 250\,Hz \cite{7989345}. There are total around 6,000 experimental cases and
these data are open to download from \cite{MCubeLab_PushData}.

\begin{table}[ht]
\caption{Push Data Cases}
\label{push_data_cases}
\begin{center}
\begin{tabular}{c||c|c}
\hline
	Case \# & acceleration [m/s$^2$] & velocity [m/s]\\
\hline
	1 & 0.0 & 0.01\\
	2 & 0.0 & 0.01\\
	3 & 0.0 & 0.01\\
	4 & 0.0 & 0.05\\
	5 & 0.2 & -0.001\\
	6 & 2.5 & -0.001\\
\hline
\end{tabular}
\end{center}
\end{table}

\begin{table}[ht]
\caption{Push Data Measurements}
\label{push_data_measurements}
\begin{center}
\begin{tabular}{l||l}
\hline
Measurements & Definitions\\
\hline
	${\bf r}^{\{\text{rbt}\}}_\text{C}$ & the tip position in $\{\text{rbt}\}$\\
	$\theta_\text{tip}$ & the tip orientation with respect to $\{R\}$\\
	${\bf r}^{\{\text{R}\}}_\text{O'}$ & the object centre position in $\{\text{R}\}$\\
	$\theta$ & the object orientation with respect to $\{R\}$\\
	$-{\bf f}^{\{\text{rbt}\}}_\text{C}$ & the reaction force measured at the tip in $\{\text{rbt}\}$\\
	$-T^{\{\text{rbt}\}}_\text{C}$ & the reaction torque measured at the tip in $\{\text{rbt}\}$\\
\hline
\end{tabular}
\end{center}
\end{table}

In this paper, six experiments shown in Table \ref{push_data_cases} are chosen
for testing algorithms to be proposed in the next section. The experiments include zero acceleration
and non-zero acceleration of the pusher with three different velocities of the pusher. The object of
the chosen experiments called {\it rec2} is a rectangular shape, 
whose width and length are 90\,mm and 112.5\,mm, respectively,
\cite{MCubeLab_PushData, DBLP:journals/corr/YuBFR16}.
The surface type is the same for all six experiments called {\it abs} \cite{7989345}, whose friction
coefficient is between 0.13 and 0.15. The available measurements are summarised in 
Table \ref{push_data_measurements}.

\begin{figure}[ht]
	\centering
	\includegraphics[width=2.75in]{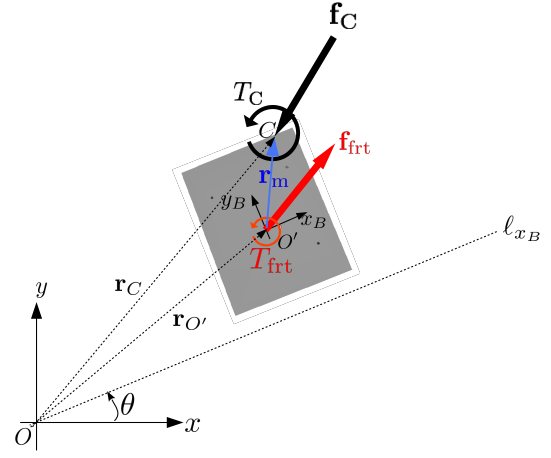}
	\caption{Push object coordinates and a free body diagram, 
		where $\ell_{x_B}$ is parallel to ${\bf x}_B$}
	\label{rec2_object}
\end{figure}

The tip position of the robot arm in the robot frame, ${\bf r}^{\{\text{rbt}\}}_\text{C}$, 
is converted into the reference frame, $\{R\}$, i.e., $x$-$y$ coordinates in Figure \ref{rec2_object}, by
\begin{align}
	{\bf r}^{\{R\}}_\text{C} = C(\theta_\text{tip}){\bf r}^{\{\text{rbt}\}}_\text{C}
\end{align}
where $\theta_\text{tip}$ is the tip orientation angle relative to the reference frame and 
the direction cosine matrix, $C(\alpha)$, is defined by
\begin{align}
C(\alpha) = 
\begin{bmatrix}
		\cos\alpha & -\sin\alpha\\
		\sin\alpha & \cos\alpha
\end{bmatrix}
\end{align}
Similarly, the force and the torque applied by the pusher tip in the reference frame is obtained by
\begin{align}
	{\bf f}^{\{R\}}_\text{C} = C(\theta_\text{tip}){\bf f}^{\{\text{rbt}\}}_\text{C}
\end{align}

\subsection{Calculating the total external force}
\begin{figure}[ht]
	\centering
	\includegraphics[width=3.65in]{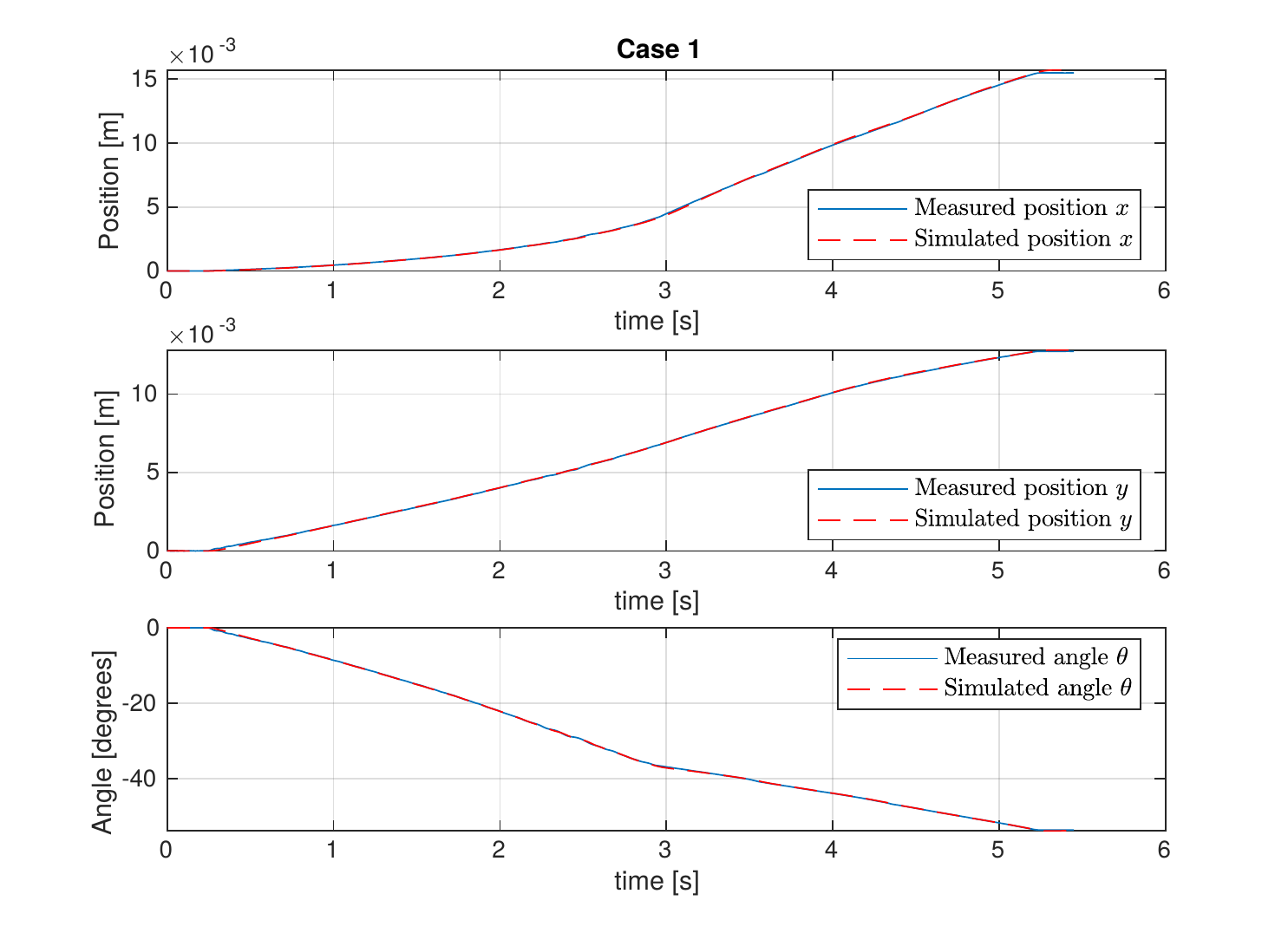}
	\caption{Position \& orientation using the estimated total force and torque}
	\label{PID_test_case_1_pos}
\end{figure}

For brevity, the superscript to indicate the coordinate frame is omitted whenever it does not 
cause any confusion.
Applying the Newton's second law of motion in the reference frame
\begin{align} \label{linear_eom}
	m\ddot{\bf r}_{\text{O}'} = {\bf f}_\text{C} + {\bf f}_\text{frt} = {\bf f}_\text{total}
\end{align}
where $m$ is the mass of the object, $\ddot{(\cdot)} = d^2(\cdot)/dt^2$, $t$ is time in seconds, 
and ${\bf f}_\text{frt}$ is the sum of all 
unknown forces including frictional forces.  The equation of motion for the rotation is 
\begin{align} \label{rotational_eom}
	I\ddot{\theta} = T_\text{C} + T_\text{frt} + {\bf r}_m \times {\bf f}_\text{C} = T_\text{total}
\end{align}
where $I$ is the moment of inertia of the object, ${\bf r}_m$ is the distance from the centre of the 
object to 
the contact point, which is difference between the tip position, ${\bf r}_\text{C}$, and 
the object centre position,
${\bf r}_{\text{O}'}$, i.e., ${\bf r}_\text{C}-{\bf r}_{\text{O}'}$, and $T_\text{frt}$ is the sum of
all unknown torque including frictional torques.

To calculate the total sum of forces, ${\bf f}_\text{total}$,
and the total sum of torques, $T_\text{total}$, which are not directly measured or provided in the
data set, a PID (Proportional-Integral-Derivative)
controller is designed for each to track the measured object position, 
${\bf r}_{\text{O}'}$, or the measured object orientation, $\theta$ as follows:
\begin{subequations}
	\begin{align}
	m\ddot{\bf r}_\text{sim} &= k_p {\bf e}_r + k_i \int_0^t{\bf e}_r(\tau)d\tau + k_d \dot{\bf e}_s
		=  \hat{\bf f}_\text{sim}\\
	I\ddot{\theta}_\text{sim} &= l_p {e}_\theta + l_i \int_0^t{e}_\theta(\tau)d\tau 
		+ l_d \dot{e}_\theta = \hat{T}_\text{sim}
	\end{align}
\end{subequations}
where $k_p$, $k_i$ and $k_d$ are the PID gains for the linear motion, 
$l_p$, $l_i$, and $l_d$ are the PID gains for the rotational motion, and
\begin{subequations}
	\begin{align}
		{\bf e}_r &= {\bf r}_{\text{O}'} - {\bf r}_\text{sim}\\
		{e}_\theta &= {\theta} - {\theta}_\text{sim}
	\end{align}
\end{subequations}
The following set of the control gains are found by trial and error:
\begin{figure}[ht]
	\centering
	\includegraphics[width=3.25in]{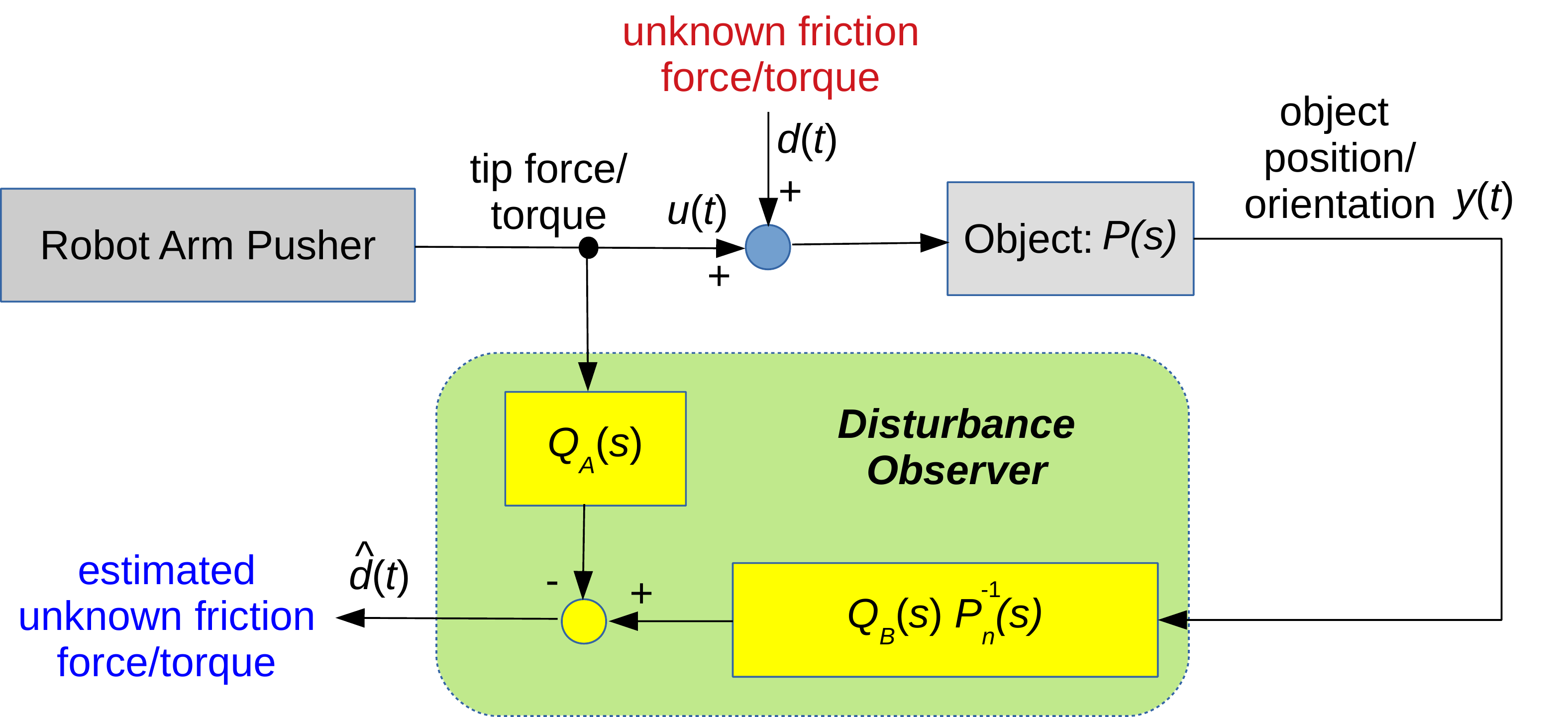}
\caption{Disturbance observer diagram: two inputs to the observer are the measured position
	or orientation of the object and the force or torque applied by the pusher.}
	\label{dob_diagram}
\end{figure}
\begin{subequations}
	\begin{align}
		k_p &= 46.6,~ k_i = 34.6,~ k_d = 15.4\\
		l_p &= 117.2,~ l_i = 137.8,~ l_d = 24.5
	\end{align}
\end{subequations}

To verify the estimated force and torques, set ${\bf f}_\text{total} =  \hat{\bf f}_\text{sim}$ 
and $T_\text{total} = \hat{T}_\text{sim}$ and solve \eqref{linear_eom} and \eqref{rotational_eom}.
The position and the orientation calculated using the estimated forces/torques are compared with the
measured position and orientation in Figure \ref{PID_test_case_1_pos} for Case 1. All trajectories
are reasonably close to the measured values. Now, the unknown force and the torque are calculated
by
\begin{subequations}
	\begin{align}
		{\bf f}_\text{frt} &\approx \hat{\bf f}_\text{sim}  - {\bf f}_\text{C}\\
		{T}_\text{frt} &\approx \hat{T}_\text{sim}  - {T}_\text{C}
	\end{align}
\end{subequations}

Although this approach provides relatively accurate force and torque estimations,
the errors in the estimation are strongly dependent on the choice of PID control gains. 
It would be very difficult to find appropriate gains working for various push scenarios.
Hence, more reliable disturbance estimation algorithm is required. The calculated 
unknown force and torque by the PID controller are to be used to confirm the performance
of disturbance observers to be designed in the next section.

\section{Algorithm}
In this section, firstly, estimating unknown forces or torques in pushing object in a planner 
surface is posed as a disturbance observer problem. Secondly, a disturbance observer is designed
and the performance of the observer is demonstrated using the push data. Finally, a prediction
algorithm is designed based on the pattern found in the estimated disturbances.

\begin{figure}[ht]
	\centering
	\includegraphics[width=3.65in]{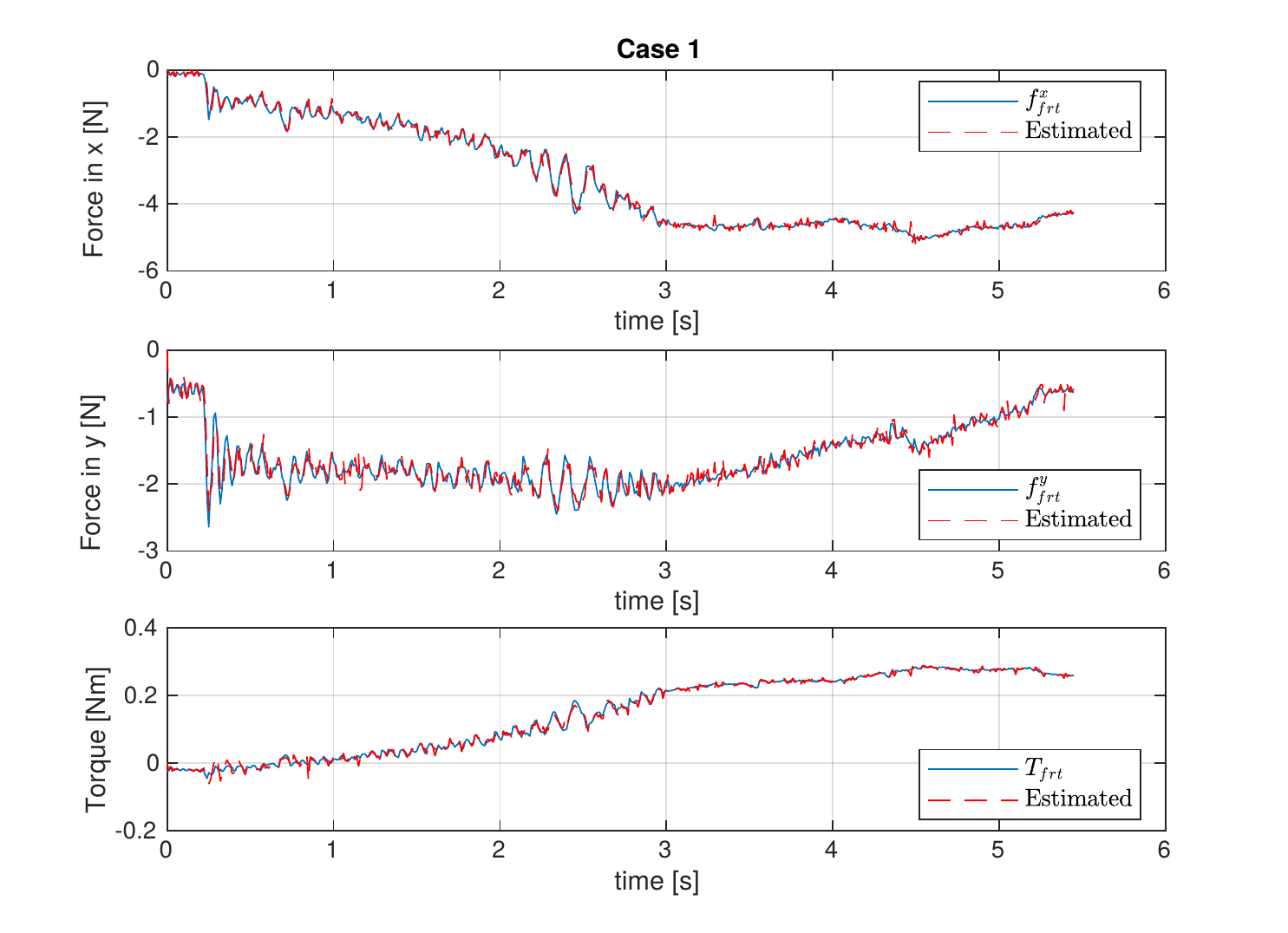}
\caption{Force \& torque estimation by the disturbance observer. The estimated disturbances 
	are closely matched with the ones from the dynamics simulations.}
	\label{DOB_force_test_case_1}
\end{figure}

\subsection{Disturbance Observer}
The Q-filter based disturbance observer is shown in Figure \ref{dob_diagram}.
The estimated total sum of the disturbances, $\hat{d}(t)$, is given by
\begin{align}
	\hat{d}(t) = - Q_A(s) u(t) + Q_B(s) P^{-1}_n(s) y(t)
\end{align}
where the Laplace domain, $s$, and the time domain, $t$, are 
used at the same time for a notational convenience, 
$u(t)$ is the pusher force or torque applied to the object, $y(t)$
is the object position or orientation measurements, and $Q_A(s)$, $Q_B(s)$ and $P_n(s)$ are
the transfer functions to be designed. Two inputs to the observer, $u(t)$ and
$y(t)$ include some measurement noise, which should be appropriately taken into account when
the observer is designed. 

If there is no sensor noise and the system transfer function, $P(s)$ is perfectly known,
then $y(t)$ is equal to $P(s) [u(t) + d(t)]$. Set $P_n(s) = P(s)$ and $Q_A(s)=Q_B(s)$,
\begin{align}
	\hat{d}(t) = Q_A(s) d(t)
\end{align}
Hence, the estimated disturbance, $\hat{d}(t)$, is equal to 
the external disturbance filtered by $Q_A(s)$. For the translational and the rotational dynamics,
\eqref{linear_eom} and \eqref{rotational_eom}, the system is a simply double integrator with the inertia,
i.e., $P(s) = 1/(m s^2)$ or $P(s) = 1/(I s^2)$. 
Only uncertain parameter for each is the inertia parameter, mass
or the moment of inertia. Mismatch between the known values and the true values can be written as
\begin{align}
	m = \bar{m} + \Delta m,~ I = \bar{I} + \Delta I
\end{align}
where $\bar{m}$ and $\bar{I}$ are the nominal values, and $\Delta m$ and $\Delta I$ are the uncertainties.
As $P_n(s) = 1/ (\bar{m} s^2)$ or $P_n(s) = 1/ (\bar{I} s^2)$, the estimated disturbance is given by
\begin{align}
	\hat{d}(t) = \alpha Q_A(s) d(t) + (1 - \alpha) Q_A(s) u(t) 
\end{align}
where $\alpha$, equal to $\bar{m}/m$ or $\bar{I}/I$, is strictly positive and becomes 1 for no uncertainty
in the inertia properties. 
For all experimental cases chosen in Table \ref{push_data_cases}, $\bar{m}$ is set
to 1.045\,kg and $\bar{I}$ is equal to 0.0018\,kg$\cdot$m$^2$ 
as they are given in \cite{MCubeLab_PushData}. The uncertainties for the inertia properties
are presumed to be very small for all the cases.

\begin{figure*}
	\subfigure[]{\includegraphics[width=2.35in]{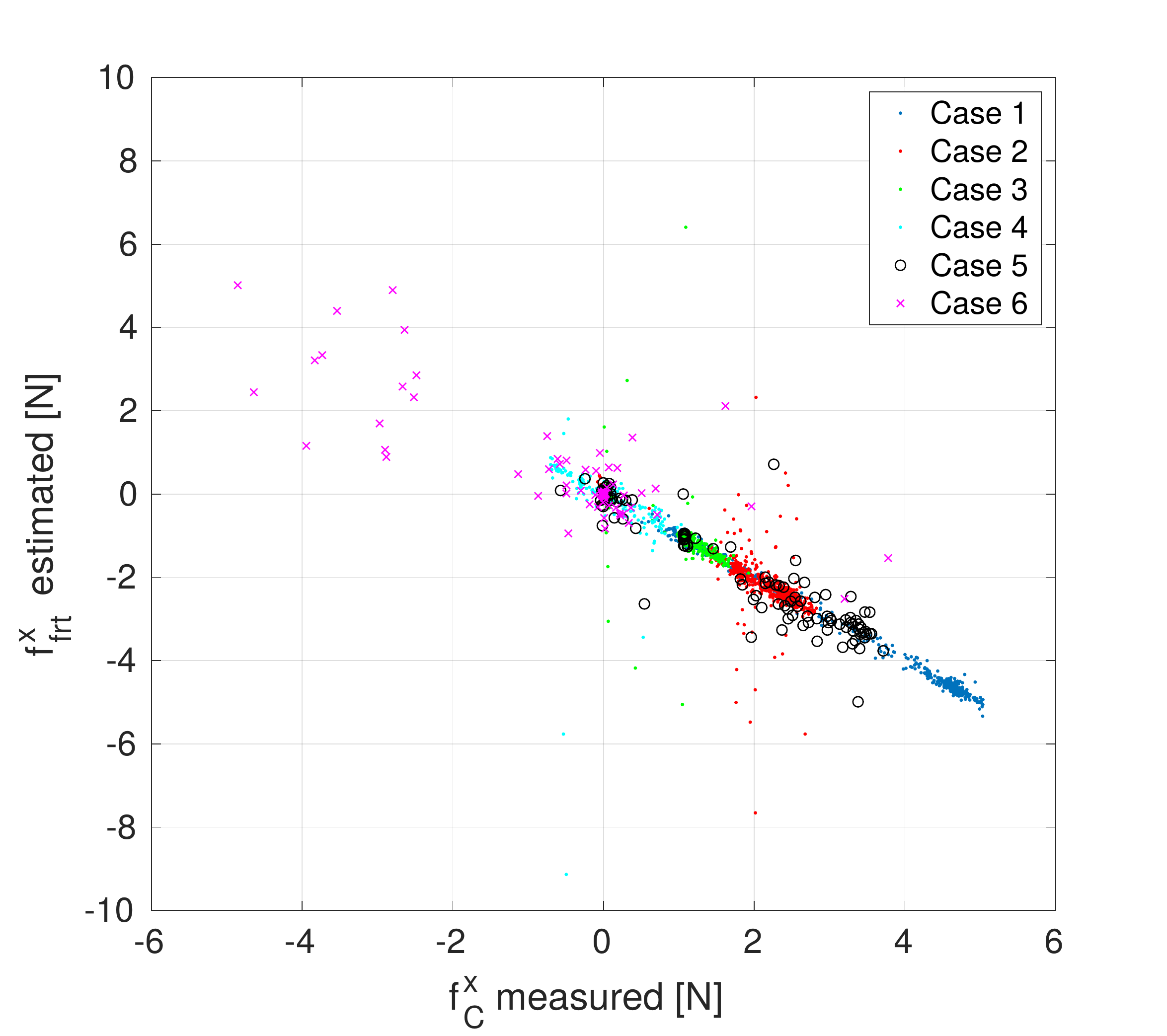}}
	\subfigure[]{\includegraphics[width=2.35in]{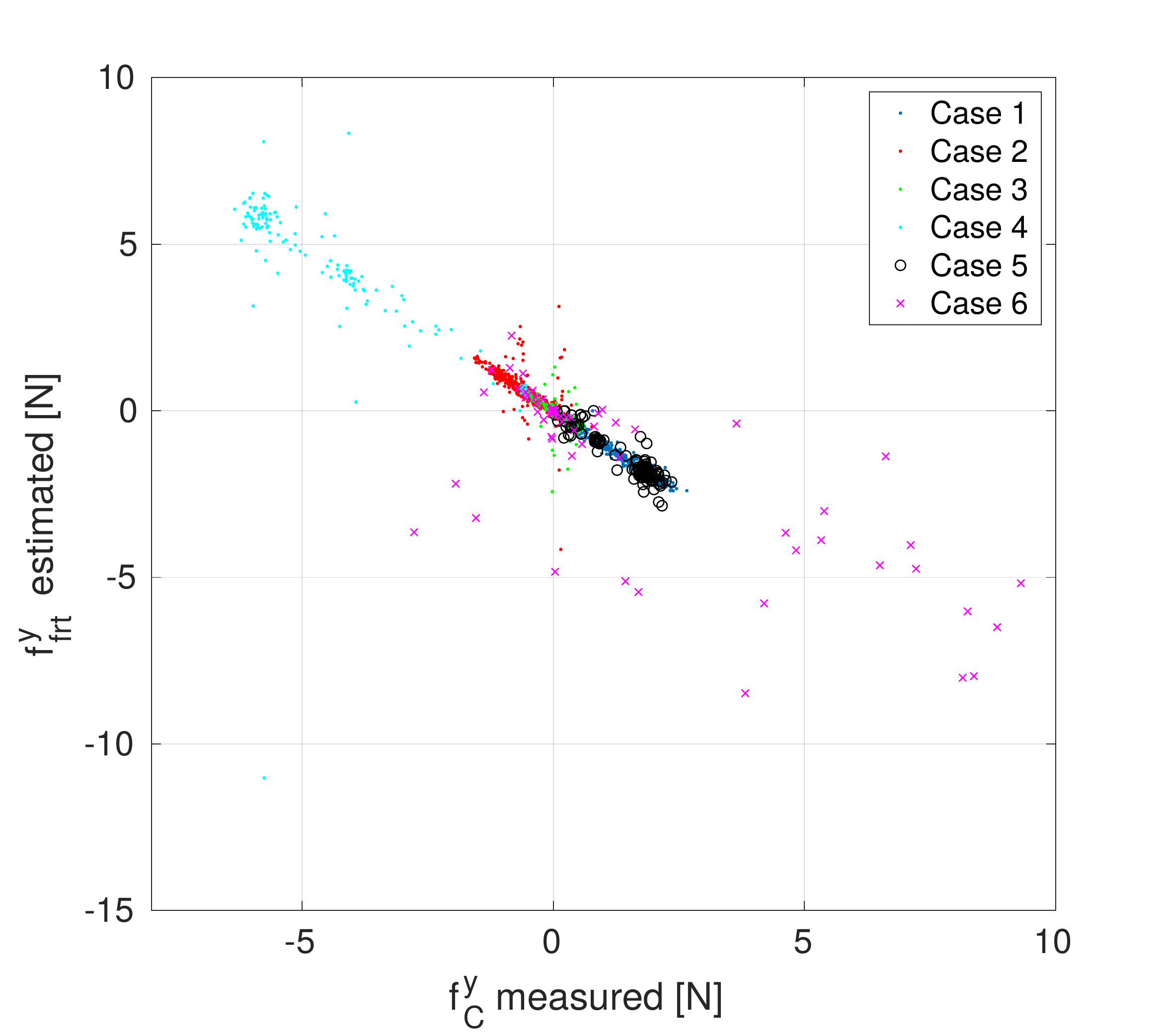}}
	\subfigure[]{\includegraphics[width=2.35in]{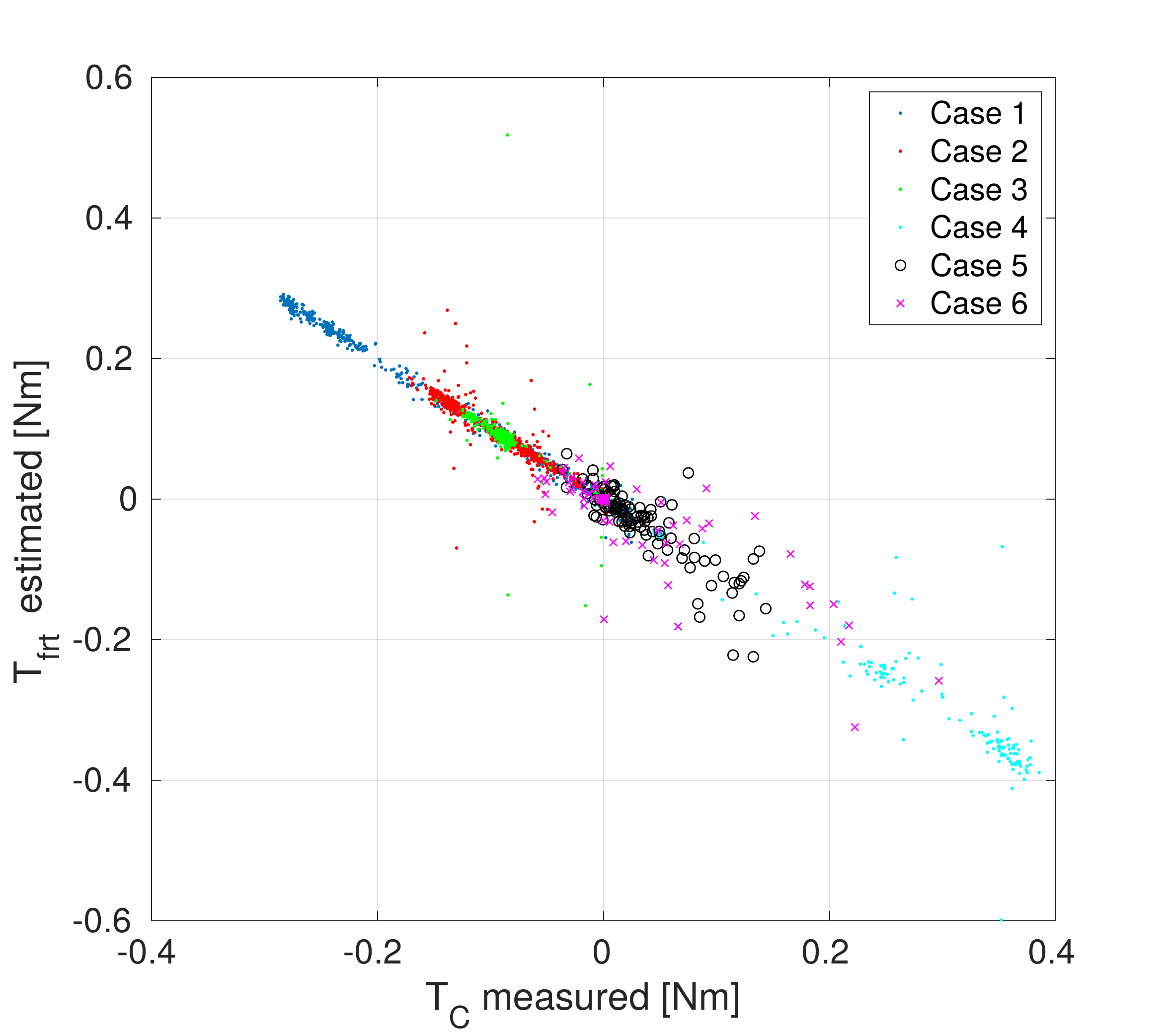}}
	\caption{The linear relationship between the measured force or torque and 
	the estimated force or torque for (a) $x$-direction, 
	(b) $y$-direction and (c) $\theta$-rotation.}
	\label{fT_meas_vs_fT_est_all}
\end{figure*}

$Q_A(s)$ and $Q_B(s)$ is set to the following form:
\begin{align} \label{Q_A_form}
	Q_A(s) = Q_B(s) = \dfrac{\omega_n^2}{s^2 + 2\zeta\omega_n s + \omega_n^2}
\end{align}
where $\zeta$ is equal to $1/\sqrt{2}$ and $\omega_n$ is equal to 300\,rad/s. The value of $\zeta$
is chosen to minimise the settling time, which is important in estimating unknown disturbances having
possibly some discontinuous jumps from time to time.
While the larger the value for $\omega_n$ covers the wider range of frequencies for the disturbances,
it would be sensitive to high frequency noises.

As shown in Figure \ref{DOB_force_test_case_1}, the disturbance observer estimates the unknown force 
or torque with very little differences from the ones calculated using the dynamic simulations with
the PID controller. It is worth to emphasise that the disturbance observer estimates the total sum
of the unknown forces or torques no matter what causes the differences. The observer estimates
current disturbances and cannot provide future disturbances.
The estimated disturbance cannot be used directly to predict a future trajectory of an object
and additional algorithm is required to predict future position and orientation of an object.

\subsection{Prediction Algorithm}
In order to use the estimated disturbances in predicting future position
and orientation of an object, a persistent pattern in estimated disturbances 
must be identified. Once the pattern is identified, future unknown forces would be predicted. 

As shown in Figure \ref{fT_meas_vs_fT_est_all}, there are clear linear 
relationships between the applied force or torque by the robot tip and the estimated unknown
disturbances by the observer. 
These linear relationships remain approximately the same for all six test cases.
The acceleration of the object seems to break the linearity as the data points
in the figures are spread away from the major cluster for the higher acceleration experiments, 
i.e., Case 5 and 6. This linear relationship can be written as
\begin{align}
	\hat{d}(t_k) = \beta u(t_k) + \epsilon
\end{align}
where $t_k \in [t_1, t_N]$ for $k=1, 2, 3, \ldots, N-1, N$, $N$ is the number of samples during 
an identification interval between $t_1$ and $t_N$, $t_1$ and $t_N$ are the start and
the end time of the identification interval, respectively,
$\hat{d}(t_k)$ is the estimated value of ${f}^x_\text{frt}$, ${f}^y_\text{frt}$ 
or $T_\text{frt}$ at time $t_k$ by the disturbance observer, 
$u(t_k)$ is the measured value of $f^x_\text{C}$, ${f}^y_\text{C}$ 
or $T_\text{C}$ at time $t_k$, 
and $\beta$ and $\epsilon$ are unknowns to be determined.\\

\noindent {\it Identification phase}: the first part of the prediction algorithm is 
finding the linear relationship.
A recursive least-square algorithm is used to determine $\beta$ and $\epsilon$ as follows 
\cite{crassidis_optimal_estimation}:
\begin{itemize}
\item {Initialise:} define the following using the current measurements and 
	the current disturbance estimation for $t_k \in [t_1, t_N]$:
\begin{subequations}
\begin{align}
	{\bf u} &= \begin{bmatrix} u(t_1) & u(t_2) & \ldots & u(t_N) \end{bmatrix}^T,\\
	{\bf d} &= \begin{bmatrix} \hat{d}(t_1) & \hat{d}(t_2) & \ldots & \hat{d}(t_N) 
	\end{bmatrix}^T,
\end{align}
\end{subequations}
where $(\cdot)^T$ is the transpose, and calculate the following:
\begin{subequations}
\begin{align}
	H_k &= \begin{bmatrix} {\bf u} & {\bf 1_N} \end{bmatrix},
		P_k^{-1} = H_k^T H_k,\\
	{\bf x}_k &= (P_k H_k^T) \hat{\bf d},
\end{align}
\end{subequations}
		where ${\bf 1}_N$ is the N$\times$1 vector whose elements are all 1, 
		${\bf x}_k = [\hat{\beta}_k~ \hat{\epsilon}_k]^T$, and $\hat{\beta}_k$
		and $\hat{\epsilon}_k$ are the current estimation of $\beta$ and
		$\epsilon$, respectively.
\item {Update:} construct $\bf{ u}$ and ${\bf d}$ using new measurement and
	estimation sets from another identification time interval, $
	t_k \in [t'_1, t'_N]$ and update each as follows:
\begin{subequations}
\begin{align}
	H_{k+1} &= \begin{bmatrix} {\bf u} & {\bf 1}_N \end{bmatrix},\\
		P_{k+1}^{-1} &= P_k^{-1} + H_{k+1}^T H_{k+1},\\
	K_{k+1} &= P_{k+1} H_{k+1}^T,\\
	{\bf x}_{k+1} &= {\bf x}_k + K_{k+1}\left( {\bf d} -H_{k+1} {\bf x}_k \right)
\end{align}
\end{subequations}
\item {Re-Initialise:}
\begin{subequations}
\begin{align}
	P_{k}^{-1} &= P_{k+1}^{-1}\\
	{\bf x}_k &= {\bf x}_{k+1}
\end{align}
\end{subequations}
When new measurement and estimation sets are available, 
go to {\it Update} phase and repeat the calculations.\\
\end{itemize}

\begin{figure}[ht]
	\centering
	\includegraphics[width=3.65in]{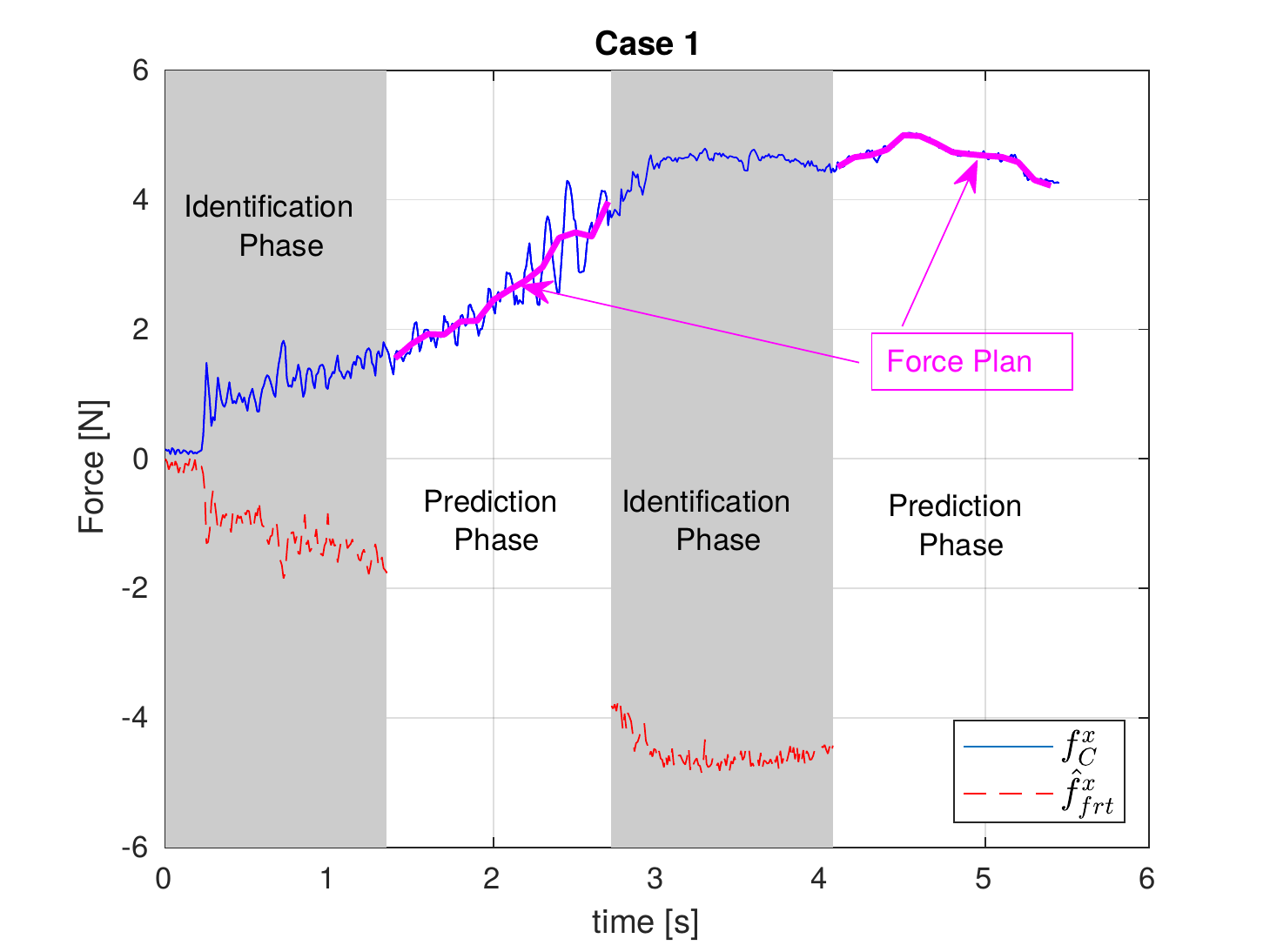}
	\caption{Prediction algorithm switching between the identification phase \& 
	the prediction phase:
	the external disturbance (red dashed lines) is estimated during the identification phase.
	For a demonstration purpose, the tip force (blue line) is sparsely sampled, 10Hz,
	for the prediction phase and it is treated
	as a force planning (pink lines) provided by a presumably high-level mission planner.}
	\label{prediction_phase_illustration}
\end{figure}

\noindent{\it Prediction Phase}: the second part of the prediction algorithm is predicting 
the position and the orientation of object using 
$\hat{\beta}_k$ and $\hat{\epsilon}_k$ found in the identification phase
To use the MCube Lab push data, the total time of the measurements are 
divided into four equal time intervals for each scenario 
as shown in Figure \ref{prediction_phase_illustration}.
The identification phase and the prediction phase are activated alternatively as shown in the figure.
To avoid using the same measured force as a planned force in the prediction phase,
where two forces might different in general, a sampled force of the measured with the frequency 10Hz,
i.e., a lot sparser force command than the actual measurements, is assumed as the planned force
given by a mission planner. Predictions of the position and the orientation are obtained by
solving the following differential equations:
\begin{subequations}
\begin{align} \label{prediction_eom}
	m\ddot{\bf r}_{\text{O}'} &= {\bf f}_\text{plan} + \text{diag}[\hat{\boldsymbol\beta}_f] 
		{\bf f}_\text{plan} + \hat{\boldsymbol\epsilon}_f\\
	I\ddot{\theta} &=  T_\text{plan} + \hat{\beta}_T T_\text{plan} + \hat{\epsilon}_T
\end{align}
\end{subequations}
where
\begin{subequations}
\begin{align}
{\bf f}_\text{plan} &= {\bf f}_C(t_k),\\
T_\text{plan} &= T_\text{C}(t_k) +  {\bf r}_m(t_k) \times {\bf f}_\text{C}(t_k),\\
\text{diag}[\hat{\boldsymbol\beta}_f] &= \begin{bmatrix}  \hat{\beta}_{fx} & 0\\ 
					0 & \hat{\beta}_{fy}\end{bmatrix},~
\hat{\boldsymbol\epsilon}_f = \begin{bmatrix} \hat{\epsilon}_{fx} & \hat{\epsilon}_{fy} 
\end{bmatrix}^T,
\end{align}
\end{subequations}
$\hat{\beta}_{fx}$ and $\hat{\epsilon}_{fx}$ are the estimated values by
the least-square algorithm for $x$-directional motions using the data collected during the identification phase,
similarly,  $\hat{\beta}_{fy}$, $\hat{\epsilon}_{fy}$,  $\hat{\beta}_{T}$, and
$\hat{\epsilon}_T$ are defined, 
the initial position and orientation are set to equal to the last measurements in the previous
identification phase, and the initial velocity and angular velocity are set to zero as they are
not measured directly.\\

\begin{figure}[ht]
	\centering
	\includegraphics[width=3.65in]{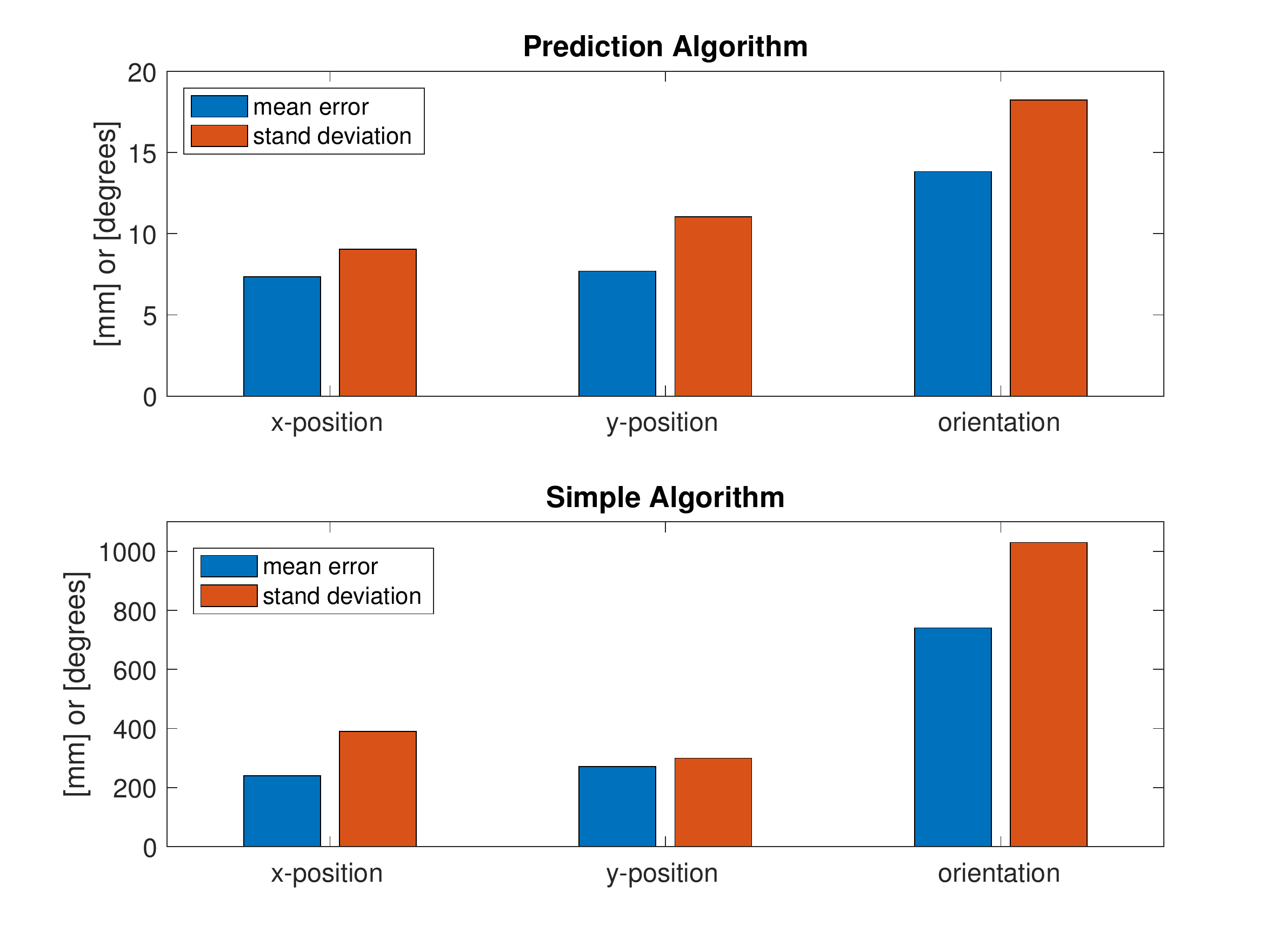}
	\caption{Average and standard deviation of the prediction errors for the proposed
	algorithm and the simple algorithm}
	\label{prediction_error}
\end{figure}

\noindent{\it Implementation}: it is important to implement the disturbance observer in
Figure \ref{dob_diagram} with $Q_A$ given in \eqref{Q_A_form} in the state-space form.
The prediction algorithm presented switches between the identification phase and 
the prediction phase instantaneously. 
Without a proper initialisation of the internal state of the disturbance observer
would cause unnecessary impulse-like peak correction at switching instances.\\

\noindent{\it Simple Algorithm}: The proposed prediction algorithm is compared with a simple 
correction algorithm, which uses a simple friction model as follows:
\begin{subequations}
\begin{align} \label{simple_prediction_eom}
	m\ddot{\bf r}_\text{O'} &= {\bf f}_\text{plan} + \mu m g\\
	I\ddot{\theta} &=  T_\text{plan} 
\end{align}
\end{subequations}
where $\mu$ is the friction coefficient, where the average value of the surface ({\it abs}), 
0.14, is used, $g$ is
the gravitation acceleration, 9.81\,m/s$^2$. Note that the rotational motion 
does not have any correction
term as it is not clear how to include a simple friction in the rotational motion.

\begin{figure}[ht]
	\centering
	\includegraphics[width=3.5in]{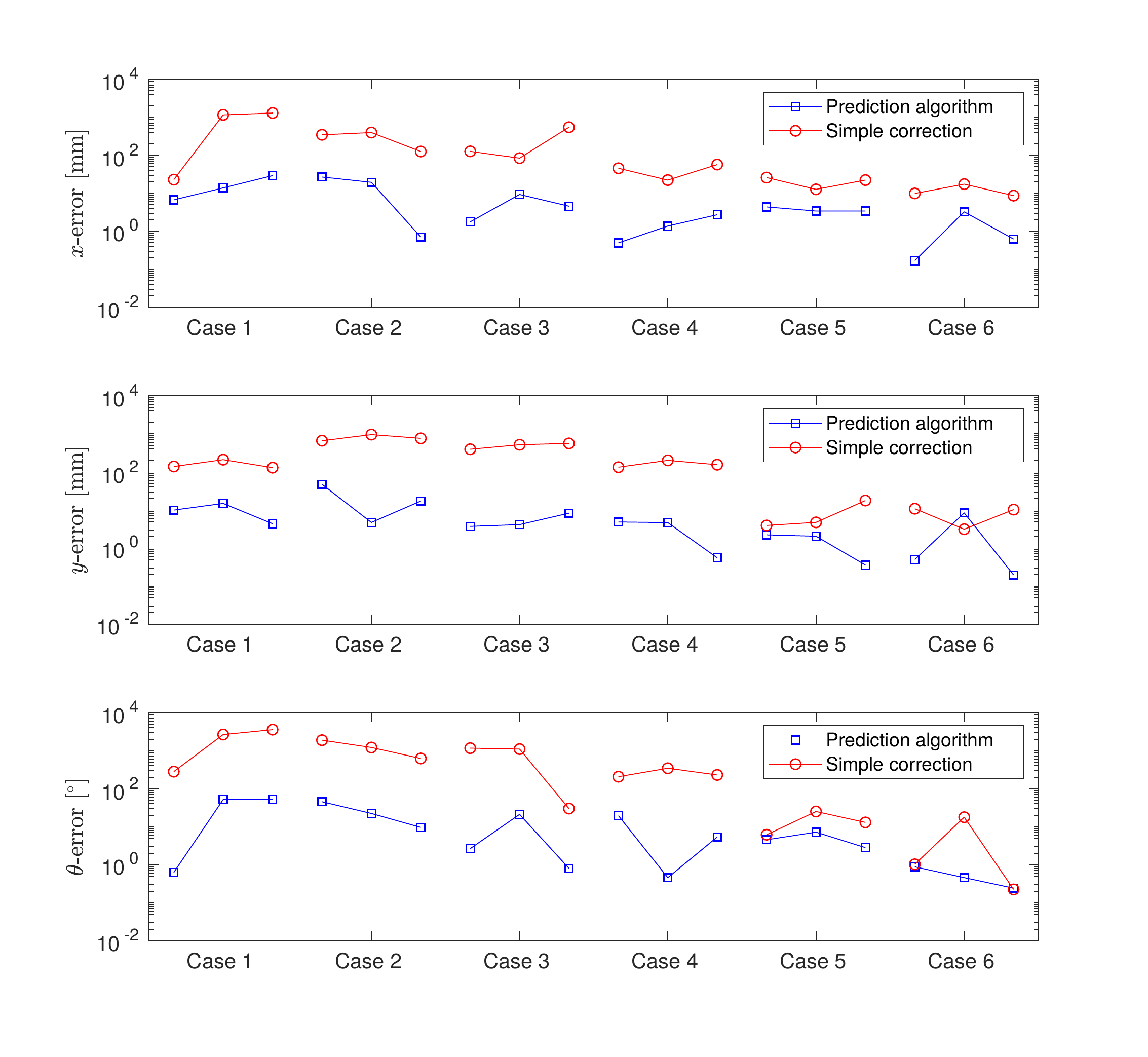}
	\caption{Prediction errors for all three prediction phases for each algorithm}
	\label{prediction_error_00}
\end{figure}

\begin{figure}[ht]
	\centering
	\includegraphics[width=3.65in]{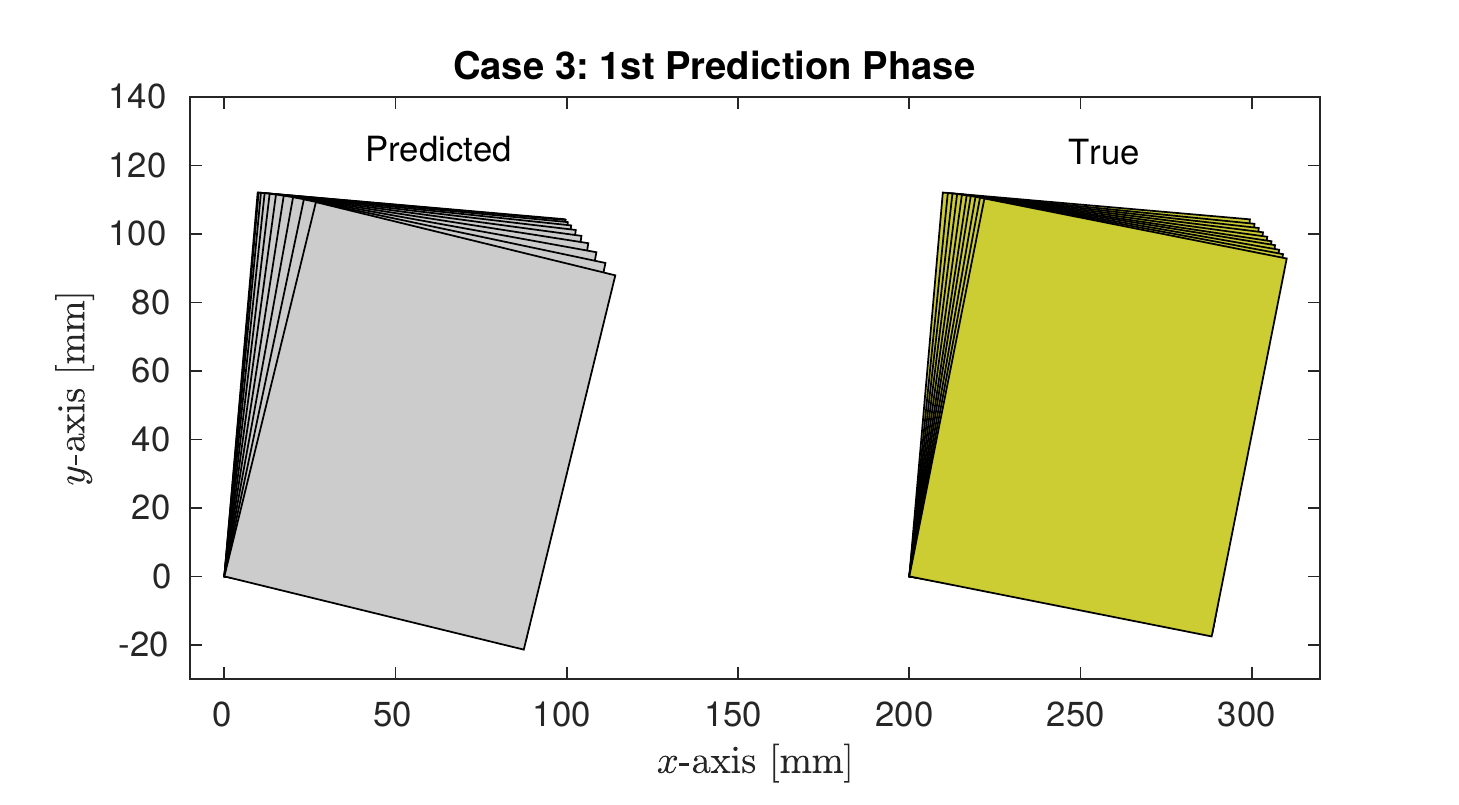}
	\caption{Best prediction case comparison}
	\label{best_prediction_case}
\end{figure}

{\it Algorithm Performance}:
The proposed prediction algorithm is tested for all six scenarios. The average error and
the standard deviation for each prediction algorithm
are shown in Figure \ref{prediction_error}. Compare to the simple correction,
the average errors for the prediction algorithm are less than 10-fold smaller, 
restricted well below 10\,mm in the $x$ and $y$ positions and 
20$^\circ$ in the orientation. 
The magnitude of error grows as the prediction horizon becomes longer.
The main reason that the errors for Case 3, 4, 5 and 6 are relatively smaller than Case 1 and 2
as shown in Figure \ref{prediction_error_00}
is their prediction horizons are shorter than the ones for Case 1 and 2 . 
The best one is the first prediction horizon for Case 3. Its object prediction is compared with the true in 
Figure \ref{best_prediction_case}.

\addtolength{\textheight}{-3.5cm}  

\section{Discussions \& Future work}
The position and orientation prediction algorithm for an object pushed by a robot in 
a planar surface is developed. 
The total sum of unknown disturbances is estimated by a disturbance observer and
a recurrent least-square algorithm estimates the correlation between 
the measured force or torque by a sensor
attached to the robot arm and the estimated disturbances by the observer. 
The correlation model, then, is used to
predict the unmodelled disturbance corresponding to a planned force or torque given by a high-level
mission planning algorithm. Finally, the position and the orientation of the object are predicted.
The performance of the algorithm is demonstrated using the push data.

The linear correlation is not, in fact, surprising for constant velocity motions as the force sum
must be zero. The relations could be a lot complex for non-zero or varying acceleration cases should
considering multiple characteristics of push motions.
For adapting to complex situations, machine vision information could be used 
with the machine learning algorithms as in \cite{Kloss2017CombiningLA}.
Also, instead of predicting a single trajectory, an ensemble of trajectories distribution described
by a probability density function would be predicted using nonlinear estimation algorithms
such as the particle filter \cite{978374} or the nonlinear projection filter \cite{7798832}.
There is also possibility to plan the robot tip movements better
for the identification phase in order to maximise information extraction about various physical
properties of object, surface, stochastic characteristics, etc.

\bibliographystyle{IEEEtran}
\bibliography{root}

\end{document}